\documentclass[a4paper, 10pt, twocolumn]{article}
\pdfoutput=1

\usepackage{amsfonts,amsmath}
\usepackage{graphicx}

\usepackage[top=1.5cm, left=1.5cm, right=1.5cm, bottom=1.5cm]{geometry}

\title{Cross-modal registration using point clouds and graph-matching in the context of correlative microscopies}

\author{Stephan Kunne$^{1}$, Guillaume Potier$^{1}$, Jean M\'{e}rot$^{1}$ and Perrine Paul-Gilloteaux$^{1,2}$\\
\footnotesize $^1$ Universit\'e de Nantes, CNRS, INSERM, l'institut du thorax, F-44000 Nantes, France \\
\footnotesize $^2$Universit\'e de Nantes, CHU Nantes, Inserm, CNRS, SFR Santé, Inserm UMS 016, CNRS UMS 3556, F-44000 Nantes, France.\
}
\date{\empty} 

\renewenvironment{abstract}{\bf\small {\em\ Abstract---}}{}

\begin{document}

\maketitle

\begin{abstract} Correlative microscopy aim at combining two or more modalities to gain more information than the one provided by one modality on the same biological structure. Registration is needed at different steps of correlative microscopies workflows. Biologists want to select the image content used for registration not to introduce bias in the correlation of unknown structures. Intensity-based methods might not allow this selection and might be too slow when the images are very large. We propose an approach based on point clouds created from selected content by the biologist. These point clouds may be prone to big differences in densities but also missing parts and outliers. In this paper we present a method of registration for point clouds based on graph building and graph matching, and compare the method to iterative closest point based methods.
\end{abstract}

\vspace*{-0.8em}
\section{Introduction}
\label{sec:introduction}

Correlative microscopies allow to combine different scales of observations and different contents, functional and morphological, based on the large panel of microscopy technologies available for life sciences \cite{walter:correlated} by registering datasets from different modalities together. The difference in scale of orders of magnitude, the level of accuracy required in biological questions, and the potential need of registration during the acquisition process call for automatic, robust and fast registration algorithm. Several approaches based on intensity matching have been proposed in the litterature based on specific metrics~\cite{acosta:intensity}, potentially constructed using deep learning~\cite{wu:features,simonovsky:crossmodalfeatures,cheng:deepsimilarity}, or using image analogies \cite{hertzmann:analogies,cao:analogies}. One potential issue with intensity-based approaches is introducing bias by using the region of interest itself in the process of registration, and then assuming correlation in this region of interest. In addition, iterative approaches based on image data volumes of several gigabytes may be too heavy in computational and memory cost.
For these reasons, we want to develop a generalized approach in 2 steps: targeted extraction of elements that are known by the biologist to be visible in both modalities; automatic matching of these elements to compute the transformation linking the targeted elements, that can then be applied to the dataset in ways less costly in memory than transforming the big image volumes. We used a point cloud based approach, where points from shapes or fiducials are extracted according to the biological questions and assumptions. In this work we focus on a point cloud registration method based on a paper by Huang et al.~\cite{huang:kinect}, and using algorithms from Papon et al.~\cite{papon:supervoxels} and Wohlkinger and Vincze~\cite{wohlkinger:esf}. We compare our method to the classically used algorithm for point cloud registration ICP/Ransac~\cite{fischler:ransac,besl:icp,chen:icp}.

\vspace*{-0.8em}
\section{Proposed registration algorithm}
\label{sec:skeleton}
\subsection{Outline}
The algorithm proceeds as follow: (1) We use a triangulation method to distribute points over surfaces that are detected in the 3d images. Then, (2-6) we follow a workflow proposed by Huang et al.~\cite{huang:kinect} to register two point clouds: (2) We build a graph for each point cloud. The graph vertices and edges represent the global geometric structure of the point cloud; hopefully, this structure is robust to the possible irregularities or differences in the two point clouds, yet carries enough information to represent the biological objects. (3) We add features on the vertices and edges of the graph, to represent the local structure of the point clouds. (4) We solve a graph matching optimisation problem, which takes into account the structure of the graph as well as the features. (5) After the graph matching, we remove outliers, \emph{i.e.}, vertices that could not be satisfyingly matched. (6) Finally, we use the resulting mapping to find a geometric transform between the two sets of vertices, and hence between the two point clouds and the two images. For steps 2, 3 and 5, we rely on the PCL library~\cite{rusu:pclib} for the implementation.

\subsection{Step by step}
\label{sec:algorithms}
\begin{figure}
\minipage{0.15\textwidth}
  \includegraphics[width=\linewidth]{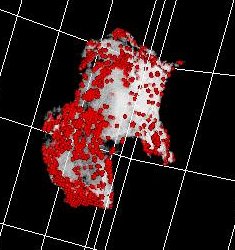}
 (a)
\endminipage\hfill
\minipage{0.15\textwidth}
  \includegraphics[width=\linewidth]{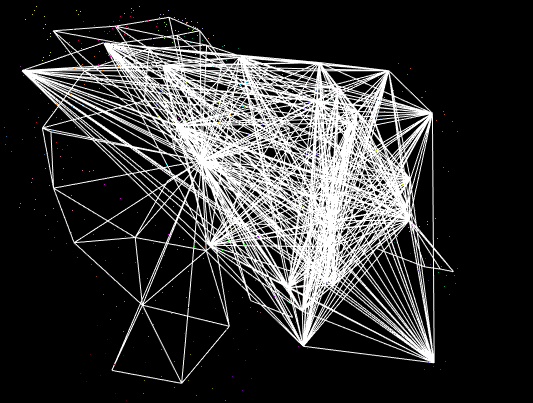}
   (b)
\endminipage\hfill
\minipage{0.15\textwidth}%
  \includegraphics[width=\linewidth]{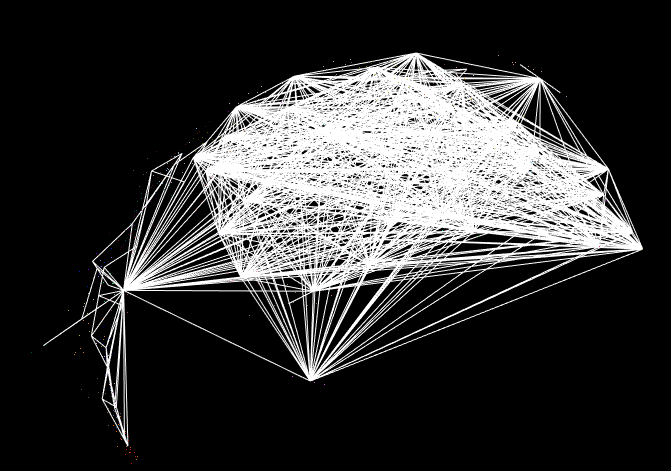}
   (c)
\endminipage
\caption{Example of data and generated graphs. (a) segmented valve extracted from microCT with points generated on its surface. (b) graph generated from this point cloud. (c) graph generated on a resampled and rotated version of the same point cloud.}
\label{fig:valve}
\end{figure}

\vspace*{-0.5em}
\paragraph{Generating point clouds.}
In our setup, point clouds are generated either from the localization of specific landmarks in the image datasets (such as spot detections for beads used as landmarks) or from the Delaunay triangulation of an ad-hoc segmentation of common elements in the images (such as nuclei, cell or organs surface,...) using the methods described in Paul-Gilloteaux et al.~\cite{paul-gilloteaux:ec-clem}. In this paper we focus on the latter (Fig.~\ref{fig:valve}).

\vspace*{-0.5em}
\paragraph{Clustering on a point cloud and building a graph.}
Following the recommendation of Huang et al.~\cite{huang:kinect}, we use supervoxel clustering~\cite{papon:supervoxels} to build a graph from a point cloud. This algorithm is implemented by J. Papon as part of the PCL library~\cite{rusu:pclib}. It consists in growing clusters of points from a set of seeds. The local clustering somewhat resembles the k-means algorithm. The centres of the clusters will become the vertices of the graph; two vertices are linked by an edge if points in the corresponding clusters are sufficiently close. An important parameter of this algorithm is the seeding resolution, which directly impacts the lengths of the edges in the resulting graph. This parameter was selected based on the sampling resolution used to create the two point clouds.

To represent accurately the structure of the whole image, the graph must be connected. We add one vertex per connected component of the graph generated by Papon's algorithm, and connect all these extra vertices.

\vspace*{-0.5em}
\paragraph{Adding features on the vertices and edges of the graph.}
During the clustering phase at the global level, a lot of information about the local structure of the point cloud was lost. Did the points in a cluster form a surface, or a volume? If it was a surface, what was its curvature? We summarize this information in the form of features on the graph vertices. Likewise, there should be features on the edges to recall the geometric structure of the graph in 3d space. For the features on the vertices, we use the "Ensemble of Shape Functions" (ESF) descriptors, originally designed to be applied to a whole point cloud to match it against a database of known shapes, in the context of 3d image classification~\cite{wohlkinger:esf}. Here, we calculate the ESF descriptor of every cluster of points, and add it as a vertex feature on the graph. We use the implementation of ESF written by Wohlkinger for the PCL library~\cite{rusu:pclib}.

\vspace*{-0.5em}
\paragraph{Graph matching.}

The problem of graph matching between two graphs $G$ and $H$ consists in finding a permutation $P$ that maps the vertices of $G$ to the vertices of $H$ in the most satisfying way. There are three criteria for satisfaction:
(1) if $(u,v)$ is an edge in $G$, then $(P(u), P(v))$ should be an edge in $H$, and reciprocally; (2) the feature of a vertex $u$ in $G$ should be as close as possible to the feature of vertex $P(u)$ in $H$; (3) the feature of an edge $(u,v)$ in $G$ should be as close as possible to the feature of edge $(P(u), P(v))$ in $H$.

A permutation $P$ can be represented by a matrix $X = (x_{i_G,i_H})^{0\leq i_G < n}_{0 \leq i_H < m}$ of dimensions $n \times m$, where $n$ and $m$ are the numbers of vertices of $G$ and $H$, respectively, and $x_{i_G,i_H} = 1$ if and only if $P(i_G)=i_H$. The problem can then be expressed as a quadratic integer programming problem, \emph{i.e.}, an optimisation problem with linear constraints and quadratic objective function:
\begin{align}
  \text{maximize: } & f(X) = X^TDX \label{eq:obj}\\
  \text{under constraints: } & \notag\\
  \forall\, 0\leq i_G < n,~ &\sum_{i_H = 0}^{m-1} x_{i_G,i_H}  \leq 1  \label{eq:rows}\\
  \forall\, 0\leq i_H < m,~ &\sum_{i_G = 0}^{n-1} x_{i_G,i_H}  \leq 1  \label{eq:cols}\\
  \forall\, i_G,i_H,~ &x_{i_G,i_H} \in \{0,1\}  \label{eq:integers}
\end{align}
The objective function $f$ depends on the matrix $D=(d^{i_G,i_H}_{j_G,j_H})^{0\leq i_G,j_G < n}_{0 \leq i_H,j_H < m}$ defined by:
\begin{itemize}
  \item $d^{i_G,i_H}_{i_G,i_H}=$ vertex feature similarity between $i_G$ and $i_H$;
  \item $d^{i_G,i_H}_{j_G,j_H}=$ edge feature similarity between $(i_G,j_G)$ and $(i_H,j_H)$ if they are edges in $G$ and $H$ respectively;
  \item $d^{i_G,i_H}_{j_G,j_H} = 0$ otherwise.
\end{itemize}
Thus the objective can be decomposed into two sums:
$$f(X) = \sum_{\substack{\text{vertices}\\i_G, i_H}} d^{i_G,i_H}_{i_G,i_H} x_{i_G,i_H}^2 + \hspace{-2em}\sum_{\substack{\text{edges}\\(i_G,j_G),(i_H,j_H)}} \hspace{-2em} x_{i_G,i_H} d^{i_G,i_H}_{j_G,j_H} x_{j_G,j_H}$$
where the first sum accounts for the satisfaction of the matched vertices, and the second sum accounts for the satisfaction of the matched edges. Note that if $x_{i_G,i_H} \in \{0,1\}$, then $x_{i_G,i_H}^2=x_{i_G,i_H}$; so that the first sum could be made linear if necessary. Efficient algorithms to solve continuous quadratic problems are well-known~\cite{frank:quadratic}.

We relaxed the integer constraint into a linear constraint $0 \leq x_{i_G,i_H} \leq 1$, and we tried solving this optimisation problem with two different solvers: (1) our own implementation of the Frank-Wolfe algorithm for quadratic programming~\cite{frank:quadratic}; (2) the nonlinear solver Ipopt~\cite{wachter:ipopt}. Frank-Wolfe's algorithm is based on the simplex algorithm and takes full advantage of the quadratic nature of the objective and the linear nature of the constraints. While Frank-Wolfe's algorithm is well-documented in the litterature, we could not find any satisfying implementation and had to reimplement it ourselves. In practice, we found that our implementation gets too often stuck in local optima. Ipopt's solver, on the other hand, is a general solver for nonlinear problems, and cannot exploit the quadratic and linear natures of the objective and constraints. Its execution time is much longer than Frank-Wolfe's algorithm, but it more often finds the optimal solution. With both solvers, we had the surprise to only ever find integer solutions. Although it might not be surprising that one of the optimal solutions is integral in some cases, litterature about graph-matching consistently argues that the difficulty of the problem resides in finding an integer solution~\cite{huang:kinect,zaslavskiy:graphmatching,zhou:graphmatching}. We will investigate with more test cases the behaviour of these solvers.

\vspace*{-0.5em}
\paragraph{Removing outliers.}During the graph-matching phase, some vertices were satisfyingly matched; it is likely that others couldn't, because they have no equivalent in the graph of the other image. In addition, because the graph-matching step consists in maximising a sum of positive terms, if there are such vertices in both graphs, then these vertices will be matched together. Before using the matching to compute the transformation, it is important to remove these artificially matched vertices. For this, we resort to the Random Sample Consensus algorithm~\cite{fischler:ransac} and its implementation in the PCL library~\cite{rusu:pclib}. The idea of this algorithm is to find the largest set of vertices in one graph that can be mapped to their respective images in the other graph by the same rigid transform.

\vspace*{-0.5em}
\paragraph{From matched nodes to geometric transform.}
From the matched vertex sets, we compute a rigid transformation using the Sch{\"o}nemann method~\cite{Schonemann1966} as implemented in eC-CLEM~\cite{paul-gilloteaux:ec-clem}.

\vspace*{-0.8em}
\section{Preliminary results}

\begin{figure}
\includegraphics[scale=0.2]{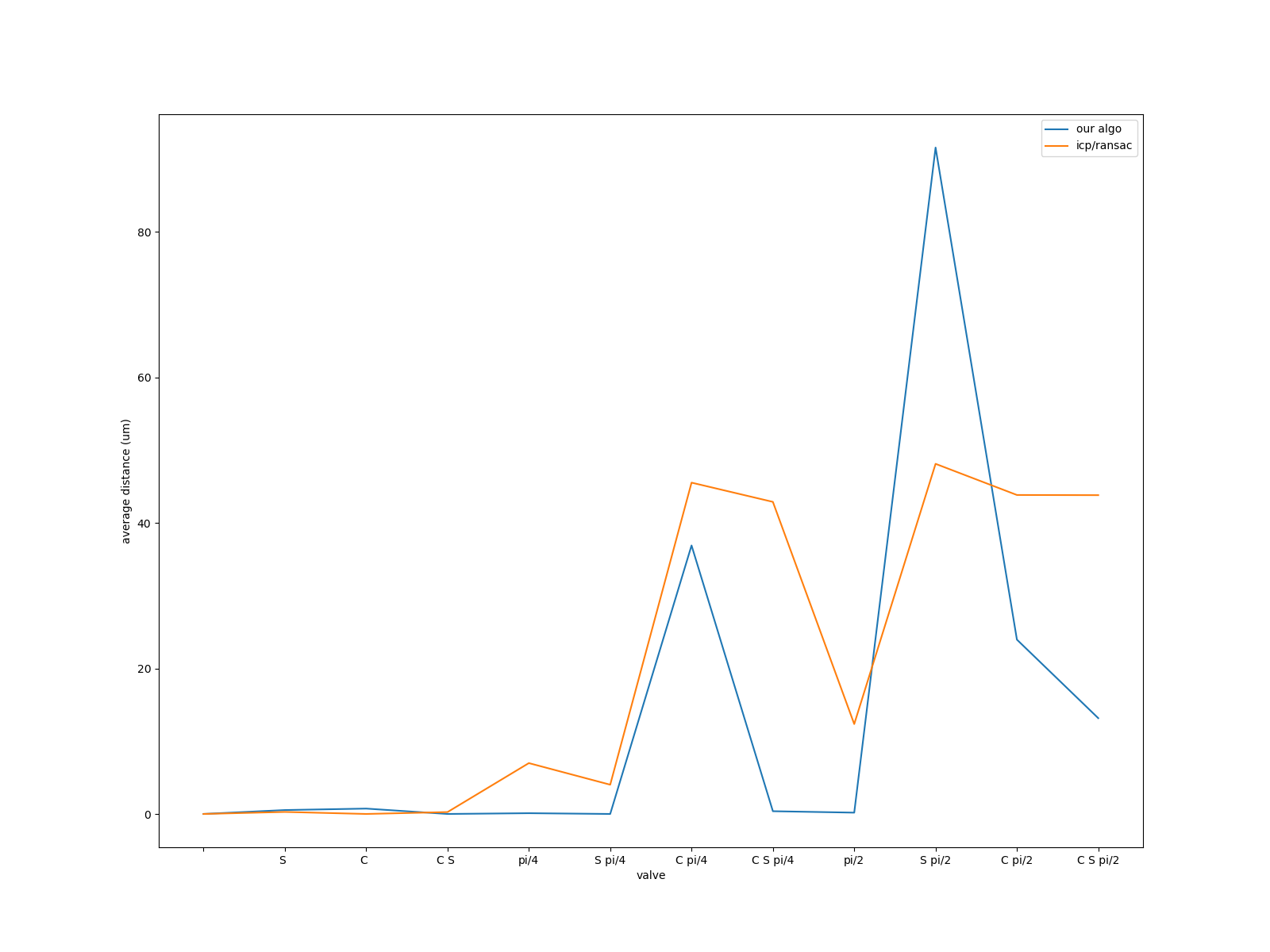}
  \caption{Mean distance between points  in micrometers in registered source point cloud and destination point cloud. The source point cloud is always the same, an artificial cloud mimicking cardiac valve. The ground truth transforms are: identity; rotation $\pi/2$ in plane $xy$. S denotes that the destination cloud is resampled by discarding $2/3$ of the points. C denotes that half of the destination cloud is cropped out.}
  \label{fig:resultsvalve}
\end{figure}

We compare our algorithm with ICP/Ransac as implemented in the PCL library. Along with a rigid transform, we simulate difference in densities and missing parts of the point cloud and apply different known transforms. We register a 3d point cloud extracted from a segmented physiological structure, a mitral valve from a rat's heart (Fig.~\ref{fig:valve}), with transformed versions of this cloud (Fig.~\ref{fig:resultsvalve}). Note that in microscopy, the voxel size is usually known such that the cloud point coordinates can be expressed in micrometers in both source and target clouds and the relation between the point cloud can be constrained to be rigid. For each pair source cloud - transformed cloud, we estimate a transformation $\hat T$ using the algorithms, then compute the average distance $\frac{1}{n}\sum_{x \text{ in source}}||\hat T(x) - T(x)||$ where $T$ is the ground truth transformation and $n$ the number of points in the source. Results presented in Fig. \ref{fig:resultsvalve} show that we outperfomed ICP/Ransac in all cases on this example.

\vspace*{-0.8em}
\section{Conclusion}
\label{sec:conclusion}
We implemented a method of registration for multimodal datasets based on point cloud registration. We have demonstrated on synthetic data obtained with simple transformations that the proposed method outperformed ICP/Ransac alone. Further simulation to represent the different cases encountered in correlated microscopies will be generated and analyzed, as well as validation on experimental data. Source code is available\footnote{https://github.com/skunne/pointcloudregistration/}.
\vspace*{-0.5em}
\paragraph{Acknowledgements}
We acknowledge ANR-18-CE45-0015.
\newpage
\bibliographystyle{ieeetr}
\bibliography{itwist20_paper}

\begin{thebibliography}{10}

\bibitem{walter:correlated}
A.~Walter, P.~Paul-Gilloteaux, B.~Plochberger, L.~Sefc, P.~Verkade,
  J.~Mannheim, P.~Slezak, A.~Unterhuber, M.~Marchetti-Deschmann, M.~Ogri,
  K.~Bühler, D.~Fixler, S.~Geyer, W.~Weninge, M.~Glösmann, S.~Handschuh, and
  T.~Wanek, ``Correlated multimodal imaging in life sciences: Expanding the
  biomedical horizon.,'' {\em Review, Front. Phys. - Medical Physics and
  Imaging}, 2020.

\bibitem{acosta:intensity}
B.~M. Toledo~Acosta, X.~Heiligenstein, G.~Malandain, and P.~Bouthemy,
  ``Intensity-based matching and registration for 3d correlative microscopy
  with large discrepancies,'' {\em IEEE Biomedical Imaging}, 2018.

\bibitem{wu:features}
G.~Wu, M.~Kim, Q.~Wang, Y.~Gao, S.~Liao, and D.~Shen, ``Unsupervised deep
  feature learning for deformable registration of {MR} brain images,'' {\em
  Medical Image Comp. and Comp.-Assisted Intervention, Nagoya, Japan}, 2013.

\bibitem{simonovsky:crossmodalfeatures}
M.~Simonovsky, B.~Guti{\'{e}}rrez{-}Becker, D.~Mateus, N.~Navab, and
  N.~Komodakis, ``A deep metric for multimodal registration,'' {\em Medical
  Image Comp. and Comp.-Assisted Intervention, Athens, Greece}, 2016.

\bibitem{cheng:deepsimilarity}
X.~Cheng, L.~Zhang, and Y.~Zheng, ``Deep similarity learning for multimodal
  medical images,'' {\em {CMBBE:} Imaging {\&} Visualization}, 2018.

\bibitem{hertzmann:analogies}
A.~Hertzmann, C.~E. Jacobs, N.~Oliver, B.~Curless, and D.~Salesin, ``Image
  analogies,'' {\em Comp. Graphics and Interactive Techniques, Los Angeles,
  USA}, 2001.

\bibitem{cao:analogies}
T.~Cao, C.~Zach, S.~Modla, D.~Powell, K.~Czymmek, and M.~Niethammer,
  ``Multi-modal registration for correlative microscopy using image
  analogies,'' {\em Medical Image Analysis}, 2014.

\bibitem{huang:kinect}
X.~Huang, J.~Zhang, L.~Fan, Q.~Wu, and C.~Yuan, ``A systematic approach for
  cross-source point cloud registration by preserving macro and micro
  structures,'' {\em IEEE Transactions on Image Processing}, 2016.

\bibitem{papon:supervoxels}
J.~Papon, A.~Abramov, M.~Schoeler, and F.~W{\"{o}}rg{\"{o}}tter, ``Voxel cloud
  connectivity segmentation - supervoxels for point clouds,'' {\em {IEEE}
  Computer Vision and Pattern Recognition, Portland, USA}, 2013.

\bibitem{wohlkinger:esf}
W.~Wohlkinger and M.~Vincze, ``Ensemble of shape functions for 3d object
  classification,'' {\em {IEEE} Robotics and Biomimetics, Karon Beach,
  Thailand}, 2011.

\bibitem{fischler:ransac}
M.~A. Fischler and R.~C. Bolles, ``Random sample consensus: a paradigm for
  model fitting with applications to image analysis and automated
  cartography,'' {\em Communications of the ACM}, 1981.

\bibitem{besl:icp}
P.~J. Besl and N.~D. McKay, ``A method for registration of 3-d shapes,'' {\em
  IEEE Transactions on Pattern Analysis and Machine Intelligence}, 1992.

\bibitem{chen:icp}
Y.~Chen and G.~G. Medioni, ``Object modeling by registration of multiple range
  images,'' {\em {IEEE} Robotics and Automation, Sacramento, USA}, 1991.

\bibitem{rusu:pclib}
R.~B. Rusu and S.~Cousins, ``3d is here: Point cloud library {(PCL)},'' {\em
  {IEEE} Robotics and Automation {(ICRA)}, Shanghai, China}, 2011.

\bibitem{paul-gilloteaux:ec-clem}
P.~Paul-Gilloteaux, X.~Heiligenstein, M.~Belle, M.-C. Domart, B.~Larijani,
  L.~Collinson, G.~Raposo, and J.~Salamero, ``{eC}-{CLEM}: flexible
  multidimensional registration software for correlative microscopies,'' {\em
  Nature Methods}, 2017.

\bibitem{frank:quadratic}
M.~Frank and P.~Wolfe, ``An algorithm for quadratic programming,'' {\em Naval
  research logistics quarterly}, 1956.

\bibitem{wachter:ipopt}
A.~W{\"{a}}chter and L.~T. Biegler, ``On the implementation of an
  interior-point filter line-search algorithm for large-scale nonlinear
  programming,'' {\em Math. Program.}, 2006.

\bibitem{zaslavskiy:graphmatching}
M.~Zaslavskiy, F.~Bach, and J.-P. Vert, ``A path following algorithm for the
  graph matching problem,'' {\em IEEE Transactions on Pattern Analysis and
  Machine Intelligence}, 2008.

\bibitem{zhou:graphmatching}
F.~Zhou and F.~De~la Torre, ``Deformable graph matching,'' {\em IEEE Conference
  on Computer Vision and Pattern Recognition, Portland, USA}, 2013.

\bibitem{Schonemann1966}
P.~H. Sch{\"o}nemann, ``A generalized solution of the orthogonal procrustes
  problem,'' {\em Psychometrika}, 1966.

\end{thebibliography}

\end{document}